\pdfoutput=1
\documentclass{bmvc2k}

\usepackage{graphicx}
\usepackage{subfigure}

\title{Deep Retinex Decomposition for Low-Light Enhancement}

\addauthor{Chen Wei $^{*}$}{weichen582@pku.edu.cn}{1}
\addauthor{Wenjing Wang $^{*}$}{daooshee@pku.edu.cn}{1}
\addauthor{Wenhan Yang}{yangwenhan@pku.edu.cn}{1}
\addauthor{Jiaying Liu $^{\dag}$}{liujiaying@pku.edu.cn}{1}

\addinstitution{
 Institute of Computer Science and Technology\\
 Peking University\\
 Beijing, P.R. China
}
\runninghead{Wei, Wang, Yang, Liu}{Deep Retinex Decomposition}

\begin{document}

\maketitle

\begin{abstract}
Retinex model is an effective tool for low-light image enhancement. It assumes that observed images can be decomposed into the reflectance and illumination. Most existing Retinex-based methods have carefully designed hand-crafted constraints and parameters for this highly ill-posed decomposition, which may be limited by model capacity when applied in various scenes. In this paper, we collect a \textbf{LO}w-\textbf{L}ight dataset (LOL) containing low/normal-light image pairs and propose a deep \textbf{Retinex-Net} learned on this dataset, including a Decom-Net for decomposition and an Enhance-Net for illumination adjustment. In the training process for Decom-Net, there is no ground truth of decomposed reflectance and illumination. The network is learned with only key constraints including the consistent reflectance shared by paired low/normal-light images, and the smoothness of illumination. Based on the decomposition, subsequent lightness enhancement is conducted on illumination by an enhancement network called Enhance-Net, and for joint denoising there is a denoising operation on reflectance. The Retinex-Net is end-to-end trainable, so that the learned decomposition is by nature good for lightness adjustment. Extensive experiments demonstrate that our method not only achieves visually pleasing quality for low-light enhancement but also provides a good representation of image decomposition.
\end{abstract}

\section{Introduction}
\label{sec:intro}
Insufficient lighting in image capturing can significantly degrade the visibility of images. The lost details and low contrast not only cause unpleasant subjective feelings, but also hurt the performance of many computer vision systems which are designed for normal-light images. There are a lot of causes for insufficient lighting, such as low-light environment, limited performance of photography equipment, and inappropriate configurations for the equipment. To make the buried details visible, improve the subjective experience and usability of current computer vision systems, low-light image enhancement is demanded.

In the past decades, many researchers have devoted their attention to solving the problem of low-light image enhancement. Many techniques have been developed to improve the subjective and objective quality of low-light images. Histogram equalization (HE)~\cite{Pizer1987Adaptive} and its variants restrain the histograms of the output images to meet some constraints. De-hazing based method~\cite{Dong2011Fast} utilizes the inverse connection between the images with insufficient illumination and those in hazy environments.

Another category of low-light enhancement methods is built on Retinex theory~\cite{Land1977The}, which assumes the observed color image can be decomposed into reflectance and illumination.  Single-scale Retinex (SSR)~\cite{Jobson1997Properties} constrains the illumination map to be smooth by Gaussian filter as the early attempt. Multi-scale Retinex (MSRCR)~\cite{Jobson1997A} extends SSR with multi-scale Gaussian filters and color restoration. \cite{Wang2013Naturalness} proposes a method to preserve naturalness of illumination with lightness-order-error measure. Fu \emph{et al.}~\cite{Fu2016B} proposed to fuse multiple derivations of the initially illumination map.  SRIE~\cite{Fu2016B} estimates reflectance and illumination simultaneously using a weighted variational model. After manipulating the illumination, the target result can be restored. LIME~\cite{Guo2017LIME}, on the other hand, only estimates illumination with structure prior and uses reflection as the final enhanced results. There are also Retinex-based methods for joint low-light enhancement and noise removal~\cite{Li2017Joint,Li2018Structure}.

Although these methods may produce promising results in some cases, they still suffer from the limitation in model capacity of the decomposition for reflectance and illumination. It is difficult to design well-working constraints for image decomposition that can be applied
in various scenes. Besides, the manipulations on illumination map are also hand-crafted and the performance of these methods usually relies on careful parameter tuning.

With the rapid development of deep neural network, CNN has been widely used in low-level image processing, including super-resolution~\cite{yang2018video,fang2018blind,yang2018reference,Yang2017Deep2}, rain removal~\cite{yang2017deep, Liu_2018_CVPR, Qian_2018_CVPR} \emph{et al.} Lore \emph{et al.}~\cite{lore2017llnet} uses stacked sparse denoising auto-encoder for simultaneous low-light enhancement and noise reduction (LLNet), however the nature of low-light pictures is not taken into account.

To overcome these difficulties, we propose a data-driven Retinex decomposition method. A deep network, called as \emph{Retinex-Net}, that integrates image decomposition and the successive enhancement operations is built. First, a subnetwork, \emph{Decom-Net} is used to split the observed image into lighting-independent reflectance and structure-aware smooth illumination. The Decom-Net is learned with two constraints. First, low/normal-light images share the same reflectance. Second, the illumination map should be smooth but retain main structures, which is obtained by a structure-aware total variation loss. Then, another \emph{Enhance-Net} adjusts the illumination map to maintain consistency at large regions while tailor local distributions by multi-scale concatenation. Since noise is often louder in dark regions and even amplified by the enhancement process, denoising on reflectance is introduced. For training such a network, we build a dataset of low/normal-light image pairs from real photography and synthetic images from RAW datasets. Extensive
experiments demonstrate that our method not only achieves pleasing visual quality in low-light enhancement but also provides a good representation of image decomposition. The contributions of our work are summarized as
follows:

\begin{itemize}
\item We build a large scale dataset with paired low/normal-light images captured in real scenes. As far as we know, it is the first attempt in the low-light enhancement field.
\item We construct a deep-learning image decomposition based on Retinex model. The decomposition network is end-to-end trained with the successive low-light enhancement network, thus the framework is by nature good at light condition adjustment.
\item We propose a structure-aware total variation constraint for deep image decomposition. By mitigating the effect of total variation at the places where gradients are strong, the constraint successfully smooths the illumination map and retains the main structures.
\end{itemize}
\section{Retinex-Net for Low-Light Enhancement}
\begin{figure}[t]
	\centering
	\begin{minipage}[t]{\linewidth}
		\centering
		\includegraphics[width=12.9cm]{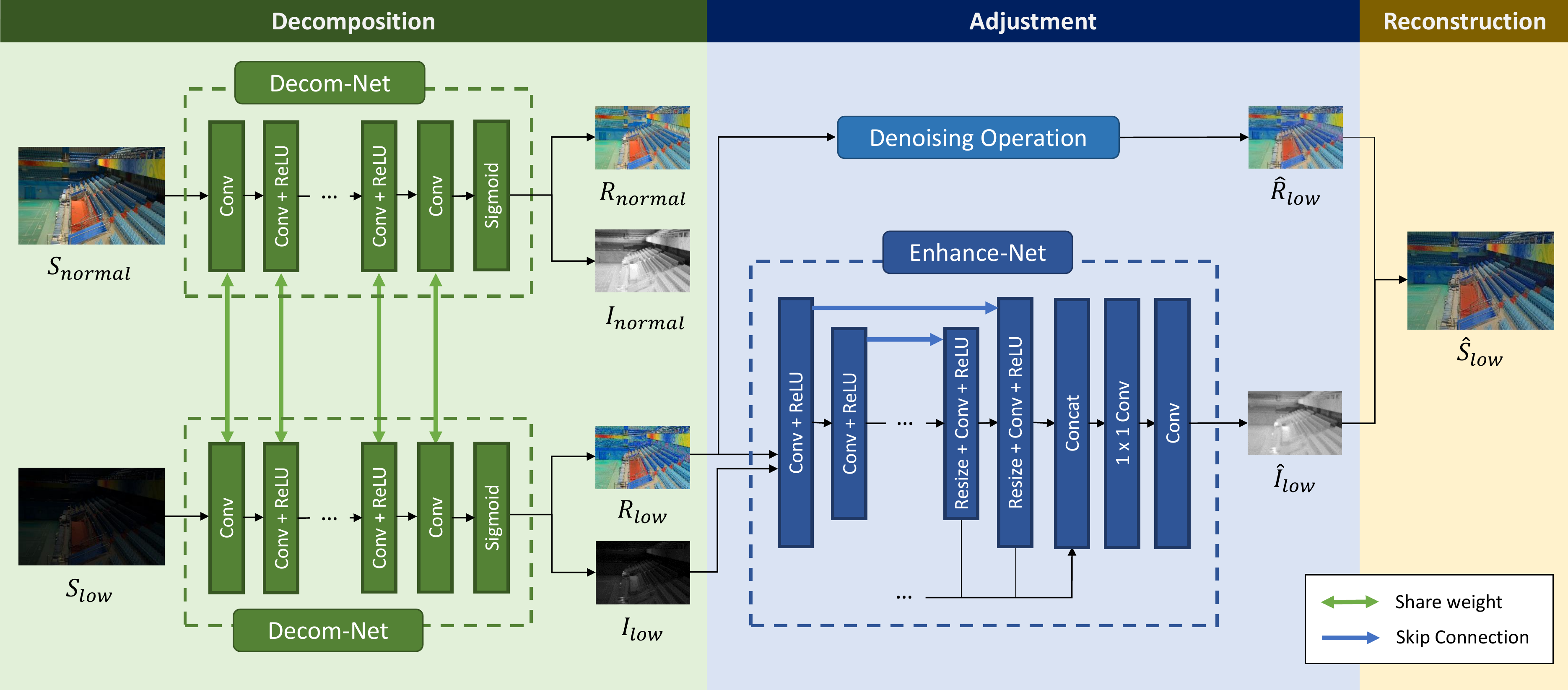}
		\end{minipage}
    \vspace{-0.6 cm}
	\caption{The proposed framework for Retinex-Net. The enhancement process is divided into three steps: decomposition, adjustment and reconstruction. In the decomposition step, a subnetwork Decom-Net decomposes the input image into reflectance and illumination. In the following adjustment step, an encoder-decoder based Enhance-Net brightens up the illumination. Multi-scale concatenation is introduced to adjust the illumination from multi-scale perspectives. Noise on the reflectance is also removed at this step. Finally, we reconstruct the adjusted illumination and reflectance to get the enhanced result.}
	\label{fig:net}
	\end{figure}

The classic Retinex theory models the human color perception. It assumes that the observed images can be decomposed into two components, reflectance and illumination. Let $S$ represent the source image, then it can be denoted by
\begin{eqnarray}
	S = R \circ I,   \label{eq:retinex}
\end{eqnarray}
where $R$ represents reflectance, $I$ represents illumination and $\circ$ represents element-wise multiplication. Reflectance describes the intrinsic property of captured objects, which is considered to be consistent under any lightness conditions. The illumination represents the various lightness on objects. On low-light images, it usually suffers from darkness and unbalanced illumination distributions.

Motivated by Retinex theory, we design a deep Retinex-Net to perform the reflectance /illumination decomposition and low-light enhancement jointly. The network consists of three steps: decomposition, adjustment, and reconstruction. At the decomposition step, Retinex-Net decomposes the input image into $R$ and $I$ by a Decom-Net. It takes in pairs of low/normal-light images at the training stage, while only low-light images as input at the testing stage. With the constraints that the low/normal-light images share the same reflectance and the smoothness of illumination, Decom-Net learns to extract the consistent $R$ between variously illuminated images in a data-driven way. At the adjustment step, an Enhance-Net is used to brighten up the illumination map. The Enhance-Net takes an overall framework of encoder-decoder. A multi-scale concatenation is used to maintain the global consistency of illumination with context information in large regions while tuning the local distributions with focused attention. Furthermore, the amplified noise, which often occurs in low-light conditions, is removed from reflectance if needed. Then, we combine the adjusted illumination and reflectance by element-wise multiplication at the reconstruction stage.

\subsection{Data-Driven Image Decomposition}
One way to decompose the observed image is estimating reflectance and illumination directly on the low-light input image with elaborately hand-crafted constraints. Since Eq.(\ref{eq:retinex}) is highly ill-posed, it is not easy to design a proper constraint function adaptive to various scenes. Therefore, we try to address this problem in a data-driven way.

During the training stage, Decom-Net takes in paired low/normal-light images each time and learns the decomposition for both low-light and its corresponding normal-light image under the guidance that the low-light image and normal-light image share the same reflectance. Note that although the decomposition is trained with paired data, it can decompose the low-light input individually in the testing phase. During training, there is no need to provide the ground truth of the reflectance and illumination. Only requisite knowledge including the consistency of reflectance and the smoothness of illumination map is embedded into the network as loss functions. Thus, the decomposition of our network is automatically learned from paired low/normal-light images, and by nature suitable for depicting the light variation among the images under different light conditions.

One thing to note is that although this problem may be similar to intrinsic image decomposition in form, they are different essentially.
In our task, we do not need to obtain the actual intrinsic image accurately, but a good representation for light adjustment.
Thus, we let the network learn to find the consistent component between low-light image and its corresponding enhanced result.


As illustrated in Fig.~\ref{fig:net}, Decom-Net takes the low-light image $S_{low}$ and the normal-light one $S_{normal}$ as input, then estimates the reflectance $R_{low}$ and the illumination $I_{low}$ for $S_{low}$, as well as $R_{normal}$ and $I_{normal}$ for $S_{normal}$, respectively. It first uses a $3\times3$ convolutional layer to extract features from the input image. Then, several $3\times3$ convolutional layers with Rectified Linear Unit (ReLU) as the activation function are followed to map the RGB image into reflectance and illumination. A $3\times3$ convolutional layer projects $R$ and $I$ from feature space, and sigmoid function is used to constrain both $R$ and $I$ in the range of [0, 1].

\begin{figure}
	\begin{minipage}[t]{\linewidth}
		\centering
		\includegraphics[width=12.9cm]{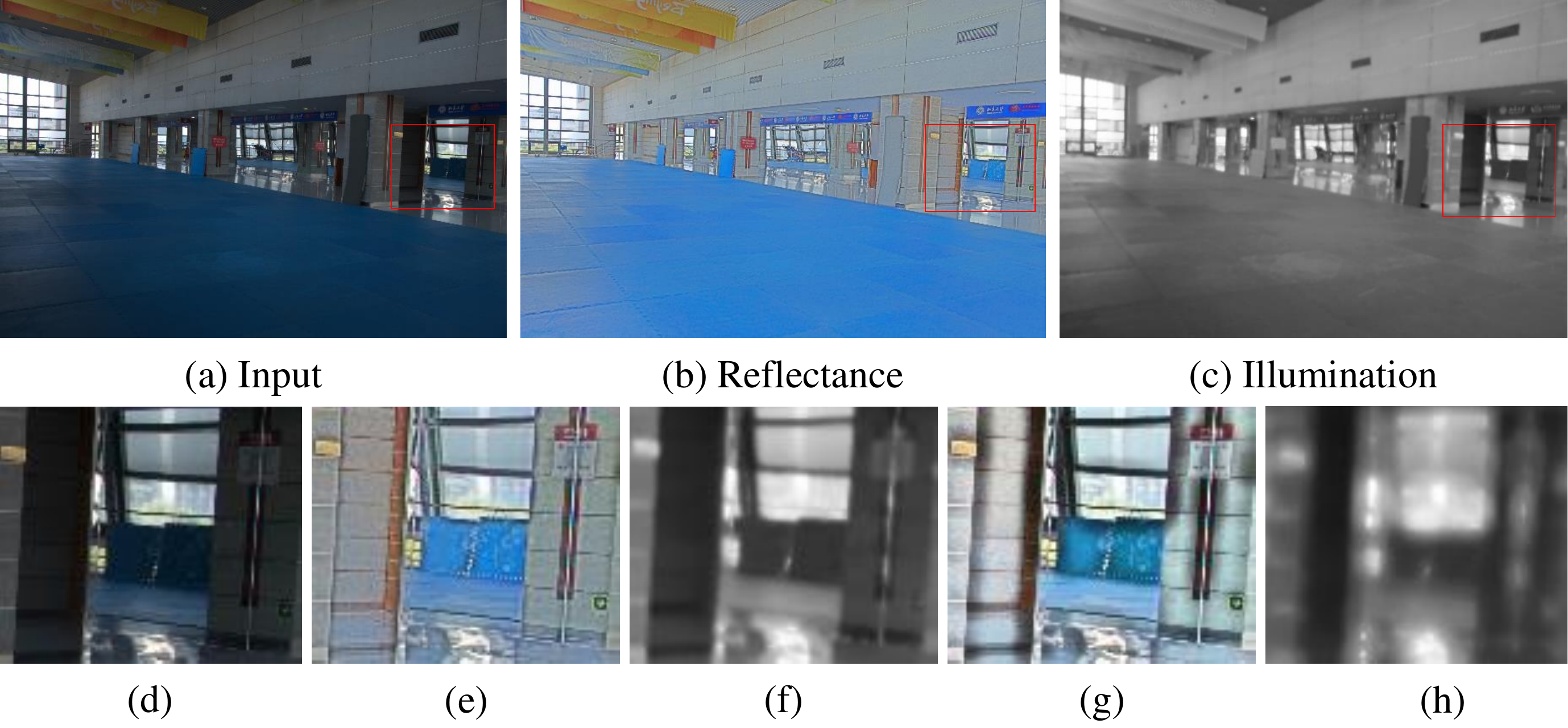}
        \vspace{-0.6cm}
	    \caption{Illustration for effectiveness of our reflectance gradient-weighted TV loss for illumination smoothness. The first row displays the input image (a), reflectance (b) and illumination (c) with weighted TV loss from left to right. The second row displays a zoom-in region where (d) is for input image, (e) and (f) are for $R$ and $I$ with weighted TV loss, (g) and (h) are for $R$ and $I$ with original TV loss.}
	    \label{fig:smoothness_loss}
    \end{minipage}
\end{figure}

The loss $\mathcal{L}$ consists of three terms: reconstruction loss $\mathcal{L}_{recon}$, invariable reflectance loss $\mathcal{L}_{ir}$, and illumination smoothness loss $\mathcal{L}_{is}$:
\begin{eqnarray}
\mathcal{L} = \mathcal{L}_{recon} + \lambda_{ir} \mathcal{L}_{ir} + \lambda_{is} \mathcal{L}_{is}, \label{eq:loss}
\end{eqnarray}
where $\lambda_{ir}$ and $\lambda_{is}$ denote the coefficients to balance the consistency of reflectance and the smoothness of illumination.

Based on the assumption that both $R_{low}$ and $R_{high}$ can reconstruct the image with the corresponding illumination map, the reconstruction loss $\mathcal{L}_{recon}$ is formulated as:
\begin{eqnarray}
\mathcal{L}_{recon} = \sum_{i=low, normal}\sum_{j=low, normal} \lambda_{ij} || R_i \circ I_j - S_j ||_1.
\end{eqnarray}
Invariable reflectance loss $\mathcal{L}_{ir}$ is introduced to constrain the consistency of reflectance:
\begin{eqnarray}
\mathcal{L}_{ir} = || R_{low} - R_{normal} ||_1.
\end{eqnarray}
Illumination smoothness loss $\mathcal{L}_{is}$ is described in detail in the following section.

\subsection{Structure-Aware Smoothness Loss}
 One basic assumption for illumination map is the local consistency and the structure- awareness, as mentioned by \cite{Guo2017LIME}. In other words, a good solution for illumination map should be smooth in textural details while can still preserve the overall structure boundary.

 Total variation minimization (TV)~\cite{Chan2011An}, which minimizes the gradient of the whole image, is often used as smoothness prior for various image restoration tasks. However, directly using TV as loss function fails at regions where the image has strong structures or where lightness changes drastically. It is due to the uniform reduction for gradient of illumination map regardless of whether the region is of textual details or strong boundaries. In other words, TV loss is structure-blindness. The illumination is blurred and strong black edges are left on reflectance, as illustrated in Fig.~\ref{fig:smoothness_loss}.

 To make the loss aware of the image structure, the original TV function is weighted with the gradient of reflectance map. The final $\mathcal{L}_{is}$ is formulated as:
\begin{eqnarray}
\mathcal{L}_{is} = \sum_{i=low, normal}|| \nabla I_i \circ exp(- \lambda_g \nabla R_i) ||, \label{eq:smoothness_loss}
\end{eqnarray}
where $\nabla$ denotes the gradient including $\nabla_h$ (horizontal) and $\nabla_v$ (vertical), and $\lambda_g$ denotes the coefficient balancing the strength of structure-awareness. With the weight $exp(- \lambda_g \nabla R_i)$, $\mathcal{L}_{is}$ loosens the constraint for smoothness where the gradient of reflectance is steep, in other words, where image structures locate and where the illumination should be discontinuous.

Although LIME~\cite{Guo2017LIME} also considers to keep image structures in illumination map with weighted TV constraint, we argue that the two methods are different. For LIME, the total variation constraint is weighted by an initial illumination map, which is the maximum intensity of each pixel in R, G and B channels. Our structure-aware smoothness loss instead is weighted by reflectance. The static initial estimation used in LIME may not depict the image structure as well as reflectance does, since reflectance is assumed as the physical property of an image. Since our Decom-Net is trained off-line with large-scale of data, the illumination and weight (the reflectance) can be updated simultaneously in training phase.

\subsection{Multi-Scale Illumination Adjustment}
The illumination enhancement network takes an overall framework of an encoder-decoder architecture. To adjust the illumination from hierarchical perspectives, we introduce a multi-scale concatenation, as shown in Fig.~\ref{fig:net}.

An encoder-decoder architecture obtains context information in large regions. The input image is successively down-sampled to a small scale, at which the network can have a perspective of the large-scale illumination distribution. This brings network the ability of adaptive adjustment. With large-scale illumination information, up-sampling blocks reconstruct local illumination distribution. Skip connections are introduced from a down-sampling block to its corresponding mirrored up-sampling block by element-wise summation, which enforces the network to learn residuals.

To adjust the illumination hierarchically, which means to maintain the consistency of global illumination while tailor the diverse local illumination distribution, a multi-scale concatenation is introduced. If there are $M$ progressively up-sampling blocks, each of which extracts a $C$ channel feature map, we resize these features at different scales by nearest-neighbor interpolation to the final scale and concatenate them to a $C \times M$ channel feature map. Then, by a $1\times1$ convolutional layer, the concatenated features are reduced to $C$ channels. A $3\times3$ convolutional layer is followed to reconstruct the illumination map $\tilde{I}$.

A down-sampling block consists of a convolutional layer with stride $2$ and a ReLU. In the up-sampling block, a resize-convolutional layer is used. As demonstrated in \cite{46191}, it can avoid checkerboard pattern of artifacts. Resize-convolutional layer consists of a nearest-neighbor interpolation operation, a convolutional layer with stride 1, and a ReLU.

The loss function $\mathcal{L}$ for Enhance-Net consists of the reconstruction loss $\mathcal{L}_{recon}$ and the illumination smoothness loss $\mathcal{L}_{is}$. $\mathcal{L}_{recon}$ means to produce a normal-light $\hat{S}$, which is
\begin{eqnarray}
\mathcal{L}_{recon} = || R_{low} \circ \hat{I} - S_{normal} ||_1.
\end{eqnarray}
$\mathcal{L}_{is}$ is the same as Eq.(\ref{eq:smoothness_loss}) except that $\hat{I}$ is weighted by gradient map of $R_{low}$.

\subsection{Denoising on Reflectance}
In the decomposition step, several constraints are imposed to the network, one of which is the structure-aware smoothness of illumination map. When the estimated illumination map is smooth, details are all retained on the reflectance, including boosted noise. Therefore, we can operate denoising method on reflectance before reconstructing the output image with illumination map. Given that noise in dark regions is amplified according to the lightness intensity during the decomposition, we should use illumination-related denoising method. Our implementation is described in Sec.~\ref{sec:experiment}.
\begin{figure}
	\centering
	\begin{minipage}[t]{\linewidth}
		\centering
		\includegraphics[width=12.9cm]{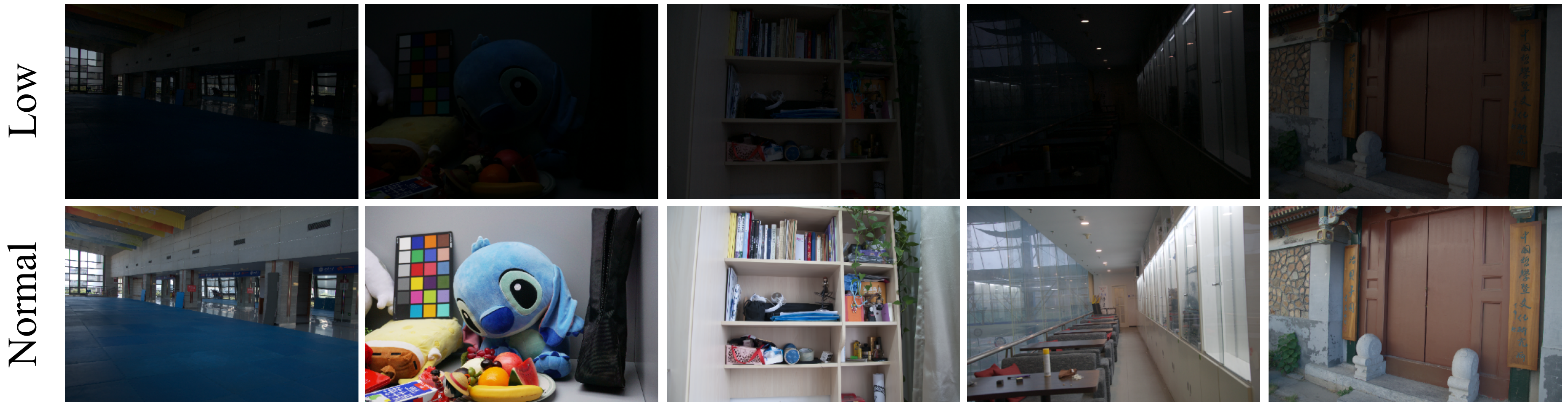}
		\end{minipage}
    \vspace{-0.6 cm}
	\caption{Several examples for low/normal-light image pairs in LOL dataset. Objects and scenes captured in this dataset are diverse.}
	\label{fig:ourdataset}
\end{figure}

\section{Dataset} \label{sec:dataset}
Although the low-light enhancement problem has been studied for decades, to the best of our knowledge, current publicly available datasets provide no paired low/normal-light images captured in real scenes. Several low-light enhancement works use datasets of High Dynamic Range (HDR) as an alternative, such as MEF dataset~\cite{Ma2015Perceptual}. However, these datasets are in small scale and contain limited scenes. Thus, they cannot be used to train a deep network. To make it tractable to learn a low-light enhancement network from a large-scale dataset, we construct a new one consisting of two categories: real photography pairs and synthetic pairs from raw images. The first one captures the degradation features and properties in real cases. The second plays a role in data augmentation, diversifying scenes and objects.

\subsection{Dataset Captured in Real Scenes}
Our dataset, named \textbf{LO}w \textbf{L}ight paired dataset (\textbf{LOL}), contains 500 low/normal-light image pairs. To the best of our knowledge, LOL is the first dataset containing image pairs taken from real scenes for low-light enhancement.

Most low-light images are collected by changing exposure time and ISO, while other configurations of the cameras are fixed. We capture images from a variety of scenes, \textit{e.g.}, houses, campuses, clubs, streets. Fig.~\ref{fig:ourdataset} shows a subset of the scenes.

Since camera shaking, object movement, and lightness changing may cause misalignment between the image pairs, inspired by \cite{Anaya2017RENOIR}, a three-step method is used to eliminate such misalignments between the image pairs in our dataset. The implementation details can be found in the supplementary file. These raw images are resized to $400 \times 600$ and converted to Portable Network Graphics format. The dataset will be available publicly.

\subsection{Synthetic Image Pairs from Raw Images}
\begin{figure}[t]
	\centering
	\begin{minipage}[t]{\linewidth}
		\centering
		\includegraphics[width=12.9cm]{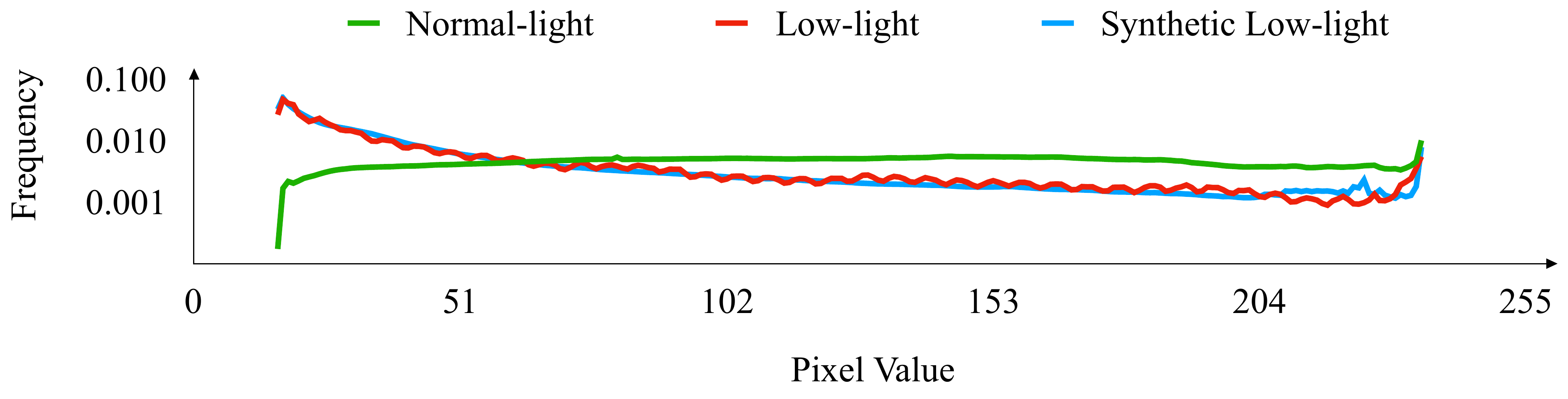}
		\end{minipage}
    \vspace{-0.6 cm}
	\caption{Fitting results based on the histogram of Y channel in YCbCr. For clarity, the histogram in depicted in the form of curve graphs and the vertical axis is scaled in logarithmic domain. The horizontal axis represents the pixel value, noticing that Y channel ranges from 16 to 240. }
	\label{fig:histogram}
	\end{figure}

To make synthetic images match the property of real dark photography, we analyze the illumination distribution of low-light images. We collect 270 low-light images from public MEF~\cite{Ma2015Perceptual}, NPE~\cite{Wang2013Naturalness}, LIME~\cite{Guo2017LIME}, DICM~\cite{Lee2013Contrast}, VV \footnote{https://sites.google.com/site/vonikakis/datasets}, and Fusion~\cite{dabov2006image} dataset, transform the images into YCbCr channel and calculate the histogram of Y channel. We also collect 1000 raw images from RAISE~\cite{Dang2015RAISE} as normal-light images and calculate the histogram of Y channel in YCbCr. Fig.~\ref{fig:histogram} shows the result.

Raw images contain more information than the converted results. When operating on raw images, all calculations used to generate pixel values are performed in one step on the base data, making the result more accurate. 1000 raw images in RAISE~\cite{Dang2015RAISE} are used to synthesize low-light images. Interface provided by Adobe Lightroom is used and we try different kinds of parameters to make the histogram of Y channel fit the result in low-light images. Final parameter configuration can be found in the supplementary material.  As shown in Fig.~\ref{fig:histogram}, the illumination distribution of synthetic images matches that of low-light images. Finally we resize these raw images to $400 \times 600$ and convert them to Portable Network Graphics format.

\section{Experiments} \label{sec:experiment}

\begin{figure}
	\begin{minipage}[t]{\linewidth}
		\centering
		\includegraphics[width=12.9cm]{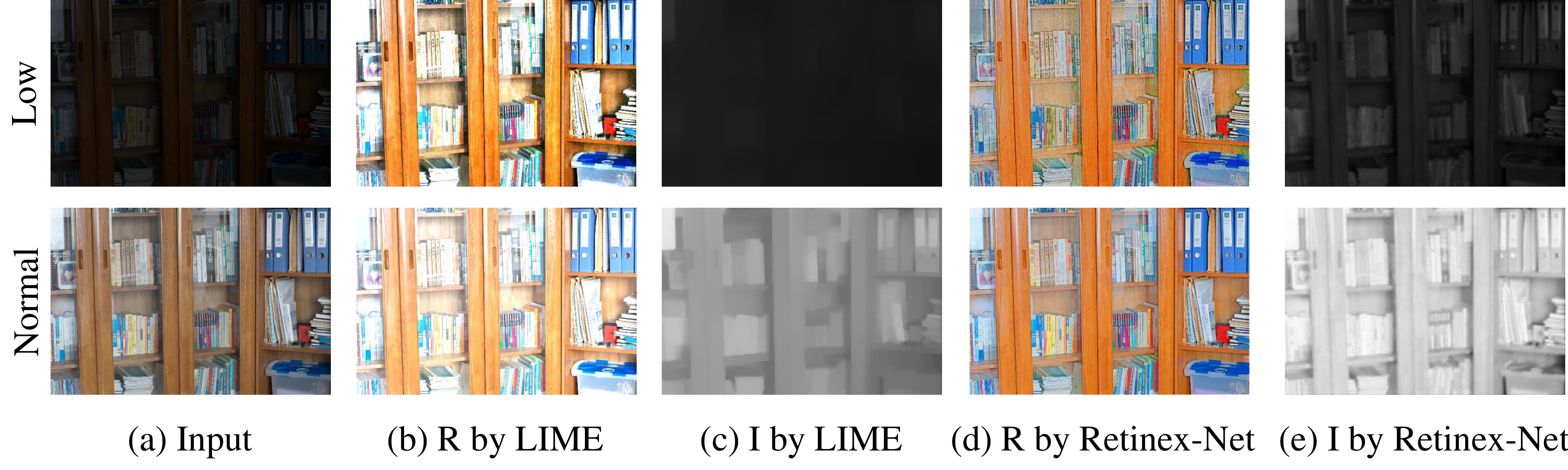}

        \vspace{-0.3cm}
	    \caption{The decomposition results using our Decom-Net and LIME on \emph{Bookshelf} in LOL dataset. In our results, the reflectance of the low-light image resembles the reflectance of the normal-light image except for the amplified noise in dark regions occurred in real scenes.}
	    \label{fig:decomposition}
    \end{minipage}
\end{figure}

\subsection{Implementation Details}
Our LOL dataset mentioned in Sec.~\ref{sec:dataset} with 500 image pairs is divided into 485 pairs for training and another 15 ones for evaluation. So the net-work is trained on 485 real-case image pairs as well as 1000 synthetic ones. The whole network is light-weighted since we empirically find it already enough for our purpose. The Decom-Net takes 5 convolutional layers with a ReLU activation between 2 conv-layers without ReLU. The Enhance-Net consists of 3 down-sampling blocks and 3 up-sampling ones. We first train the Decom-Net and the Enhance-Net, then fine-tune the network end-to-end using stochastic gradient descent (SGD) with back-propagation.  Batch size is set to be 16 and patch-size to be $96\times96$. $\lambda_{ir}$, $\lambda_{is}$ and $\lambda_g$ are set to 0.001, 0.1 and 10 respectively. When $i \not= j$, $\lambda_{ij}$ is set to 0.001, and when $ i = j $, $\lambda_{ij}$ is set to 1.
\begin{figure}[t]
	\centering
	\begin{minipage}[t]{\linewidth}
		\centering
		\includegraphics[width=12.9cm]{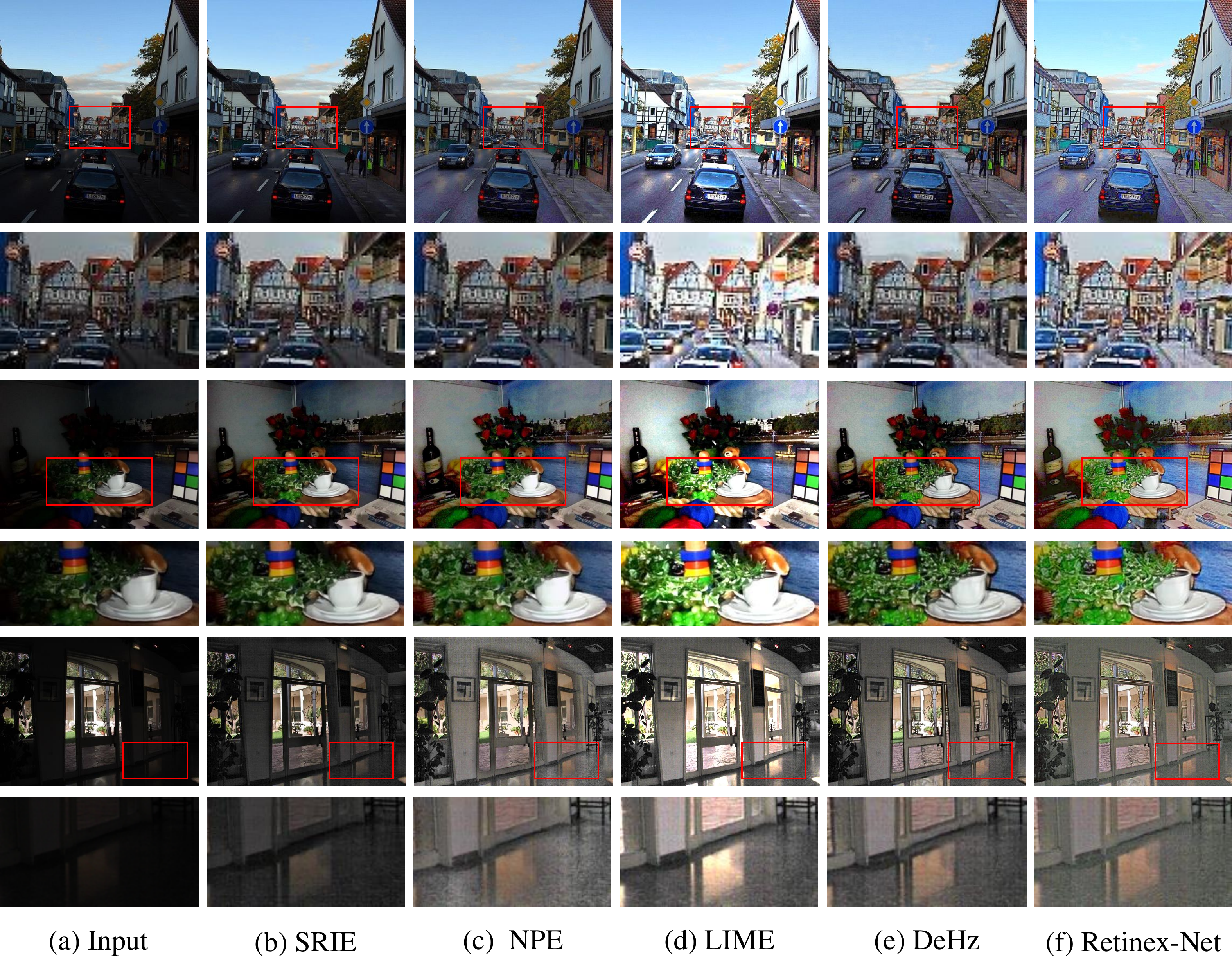}
		\end{minipage}
    \vspace{-0.6 cm}
	\caption{The results using different methods on natural images: (top-to-bottom) \emph{Street} from LIME dataset, \emph{Still lives} from LIME dataset, and \emph{Room} from MEF dataset.}
	\label{fig:compare}
\end{figure}

\begin{figure}[t]
	\centering
	\begin{minipage}[t]{\linewidth}
		\centering
		\includegraphics[width=12.9cm]{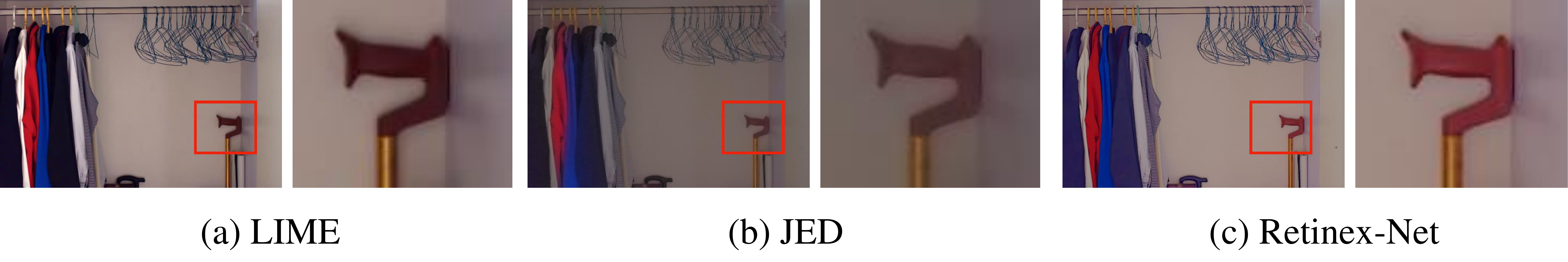}
		\end{minipage}
    \vspace{-0.6 cm}
	\caption{The joint denoising results using different methods on \emph{Wardrobe} in LOL Dataset.}
	\label{fig:denoise_evaluation}
\end{figure}

\subsection{Decomposition Results}
In Fig.~\ref{fig:decomposition} we illustrate a low/normal-light image pairs in the evaluation set of our LOL dataset, as well as the reflectance and illumination map decomposed by Decom-Net and LIME. More examples are provided in the supplementary file. It is shown that our Decom-Net can extract underlying consistent reflectance from a pair of images under quite different light conditions in both textual and smooth regions. The reflectance of the low-light image resembles the reflectance of the normal-light image except for the amplified noise in dark regions occurred in real scenes. The illumination maps, on the other hand, portray the lightness and shadow on the image. Compared with our result, LIME has much illumination information left on the reflectance (see the shadow on the shelf).

\subsection{Evaluation}
We evaluate our approach on real-scene images from public LIME~\cite{Guo2017LIME}, MEF~\cite{Ma2015Perceptual}, and DICM~\cite{Lee2013Contrast} dataset. LIME contains 10 testing images. MEF contains 17 image sequences with multiple exposure levels. DICM collected 69 images with commercial digital cameras. We compare our Retinex-Net with four state-of-the-art methods, including de-hazing based method (DeHz)~\cite{Dong2011Fast}, naturalness preserved enhancement algorithm (NPE)~\cite{Wang2013Naturalness}, simultaneous reflectance and illumination estimation algorithm (SRIE)~\cite{Fu2016A}, and illumination map estimation based (LIME)~\cite{Guo2017LIME}.

Fig.~\ref{fig:compare} shows visual comparison on three natural images. More results can be found in the supplementary file. As shown in every red rectangle, our method brightens up the objects buried in dark lightness enough without overexposure, which benefits from the learning-based image decomposition method and the multi-scale tailored illumination map. Compared with LIME, our results are not partially over-exposed (see the leaves in \emph{Still lives} and the outside leaves in \emph{Room}). The objects have no dark edges, compared with DeHz, which benefits from the weighted TV loss term (see edges on the houses in \emph{Street}).

\subsection{Joint Low-Light Enhancement and Denoising}
Considering the comprehensive performance, BM3D~\cite{dabov2006image} is used as the denoising operation in Retinex-Net. As noise is unevenly amplified on reflectance, we use a illumination relative strategy (see supplementary material). We compare our joint-denoising Retinex-Net with two methods, one is LIME with denoising post-processing, the other is JED~\cite{ren2018joint}, a recent joint low-light enhancement and denoising method. As shown in Fig.~\ref{fig:denoise_evaluation}, details are better preserved by Retinex-Net while LIME and JED blur the edges.

\section{Conclusion}
In this paper, a deep Retinex decomposition method is proposed, which can learn to decompose the observed image into reflectance and illumination in a data-driven way without the ground truth of decomposed reflectance and illumination. Subsequent light enhancement on illumination and denoising operations on reflectance are introduced. The decomposition network and low-light enhancement network are trained end-to-end. Experimental results show that our method produces visually pleasing enhancement results as well as a good representation of image decomposition.
\bibliography{egbib}
\end{document}